\documentclass[10pt,twocolumn,letterpaper]{article}

\usepackage{iccv}
\usepackage{times}
\usepackage{epsfig}
\usepackage{graphicx}
\usepackage{amsmath}
\usepackage{amssymb}
\usepackage{multirow}
\usepackage{float}
\usepackage{color}
\usepackage[accsupp]{axessibility}  

\usepackage[pagebackref=true,breaklinks=true,letterpaper=true,colorlinks,bookmarks=false]{hyperref}
\usepackage{marvosym}

\newcommand\blfootnote[1]{%
  \begingroup
  \renewcommand\thefootnote{}\footnote{#1}%
  \addtocounter{footnote}{-1}%
  \endgroup
}

\iccvfinalcopy 

\def\iccvPaperID{8355} 

\ificcvfinal\fi

\begin{document}

\title{Self Supervision to Distillation for Long-Tailed Visual Recognition}

\author{
    Tianhao Li \quad \quad Limin Wang\textsuperscript{\Letter} \quad  \quad Gangshan Wu \\
State Key Laboratory for Novel Software Technology, Nanjing University, China\\
}

\maketitle
\ificcvfinal\thispagestyle{empty}\fi

\begin{abstract}
   Deep learning has achieved remarkable progress for visual recognition on large-scale balanced datasets but still performs poorly on real-world long-tailed data. Previous methods often adopt class re-balanced training strategies to effectively alleviate the imbalance issue, but might be a risk of over-fitting tail classes. The recent decoupling method overcomes over-fitting issues by using a multi-stage training scheme, yet, it is still incapable of capturing tail class information in the feature learning stage. In this paper, we show that soft label can serve as a powerful solution to incorporate label correlation into a multi-stage training scheme for long-tailed recognition. The intrinsic relation between classes embodied by soft labels turns out to be helpful for long-tailed recognition by transferring knowledge from head to tail classes.

   Specifically, we propose a conceptually simple yet particularly effective multi-stage training scheme, termed as Self Supervised to Distillation (SSD). This scheme is composed of two parts. First, we introduce a self-distillation framework for long-tailed recognition, which can mine the label relation automatically. Second, we present a new distillation label generation module guided by self-supervision. The distilled labels integrate information from both label and data domains that can model long-tailed distribution effectively. We conduct extensive experiments and our method achieves the state-of-the-art results on three long-tailed recognition benchmarks: ImageNet-LT, CIFAR100-LT and iNaturalist 2018. Our SSD outperforms the strong LWS baseline by from $2.7\%$ to $4.5\%$ on various datasets. The code is available at \url{https://github.com/MCG-NJU/SSD-LT}.
\end{abstract}
\blfootnote{ \Letter: Corresponding author (lmwang@nju.edu.cn).}
\section{Introduction}
Deep learning has achieved remarkable progress for visual recognition in both image and video domains by training powerful neural networks on large-scale balanced and curated datasets (e.g., ImageNet~\cite{ImageNet} and Kinetics~\cite{kinetics}). Distinct from these artificially balanced datasets, real-world data always follows long-tailed distribution~\cite{longtail, weak}, which makes collecting balanced datasets more challenging, especially for classes naturally with rare samples. However, learning directly from long-tailed data induces significant performance degeneration due to the highly imbalanced data distribution.

A common series of approaches to alleviate the deterioration caused by long-tailed training data is based on class re-balanced strategies~\cite{oversample_2, costsensitive, LDAM, undersampling_1, oversampling_3}, including re-sampling training data~\cite{oversample_2, undersampling_1, featureaugmentation, oversampling_3} and designing cost-sensitive re-weighting loss functions~\cite{costsensitive, reweight_2}. These methods can effectively diminish the domination of head classes during the training procedure, and thus can yield more precise classification decision boundaries. However, they are often confronted with the risk of over-fitting tail classes since the original data distribution is distorted and over-parameterized deep networks easily fit this synthetic distribution. To overcome these issues, the recent work~\cite{decoupling, bbn} decouples the tasks of representation learning and classifier training. This two-stage training scheme first learns visual representation under the original data distribution, and then trains a linear classifier on frozen features under class-balanced sampling. This simple two-stage training scheme turns out to be able to handle the over-fitting issue and sets new state-of-the-art performance on the standard long-tailed benchmarks. Nevertheless, this two-stage training scheme fails to deal with imbalance label distribution issues well, particularly for representation learning stage.

\begin{figure*}[t]
\centering
\includegraphics[width=0.75\textwidth]{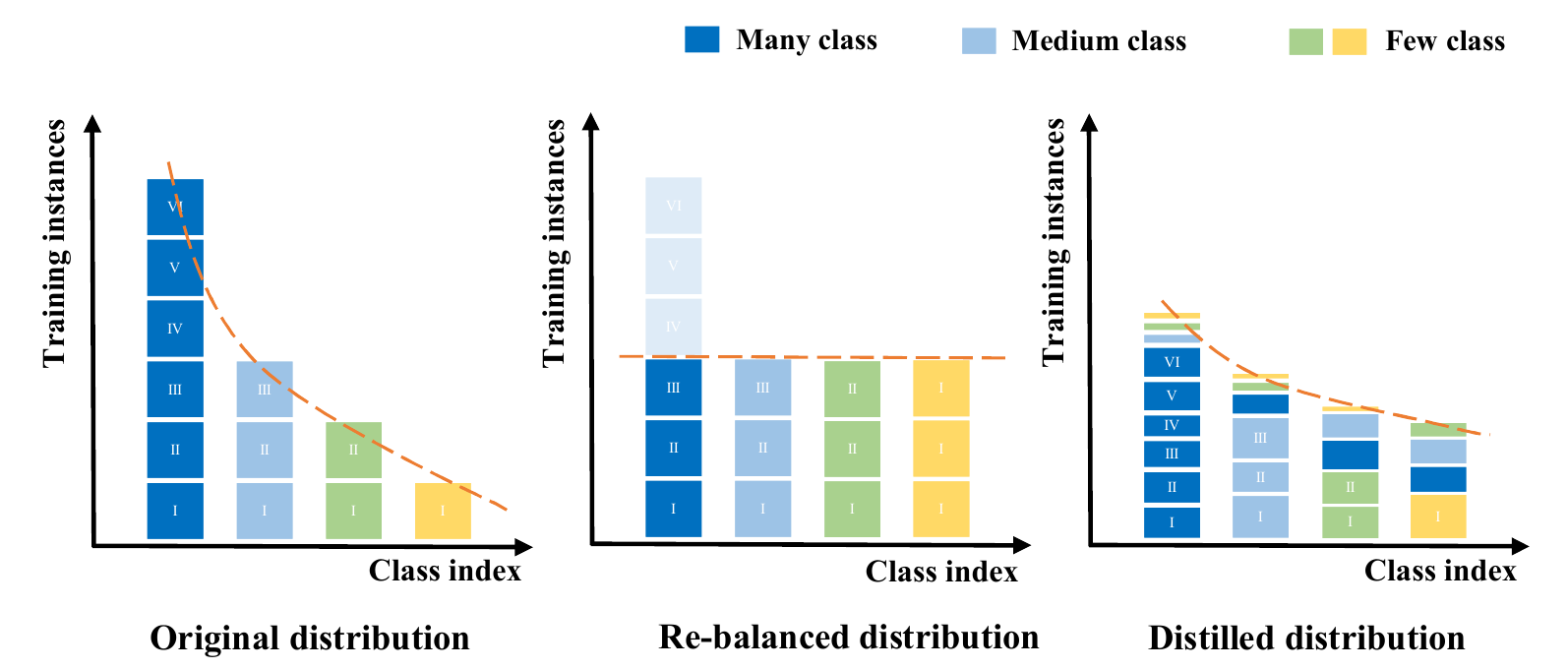}
  \caption{Real-world data always follows long-tailed data distribution, which is dominated by several head classes with abundant samples (i.e, {\color{blue}{blue}} cubes) but also contains many tail classes with scarce data (i.e. {\color{green}{green}} and {\color{yellow}{yellow}} cubes), termed as \textit{original distribution}. Learning directly from long-tailed data can cause a significant performance drop. A common way to deal with the imbalance problem is re-sampling by randomly dropping images from head classes and repeatedly sampling images from tail classes (identical image is marked by unique Roman numeral), resulting in a \textit{re-balanced distribution}. This strategy might lead to over-fitting tailed classes and under-fitting head classes. Inspired by the work of knowledge distillation in model compression, we propose to use the soft labels to deal with imbalance distribution that reflect the inherent relation between classes. The \textit{distilled distribution} acts as a naturally balanced distribution by transferring knowledge from data-rich classes to data-poor classes. Best viewed in color.}
  \label{fig:motivation}
  \vspace{-3mm}
\end{figure*}

In this paper, our objective is to design a new learning paradigm for long-tailed visual recognition, with the hope of sharing the merits of both types of long-tailed recognition methods, i.e., robust to the over-fitting issue, and effectively handling imbalance label issue. To meet this objective, our idea is to study {\em how to incorporate the label correlation into a multi-stage training scheme}? Inspired by the work of knowledge distillation~\cite{kd_hinton} in model compression, we observe that soft labels produced by a teacher network are able to capture the inherent relation between classes, which might be helpful for long-tailed recognition by transferring knowledge from head classes to tail classes, as shown in Figure~\ref{fig:motivation}. Thus, soft labels provide a practical solution for the multi-stage training strategy with label modeling.

Based on the above analysis, we present a conceptually simple yet particularly effective multi-stage training scheme for long-tailed visual recognition, termed as {\em Self Supervision to Distillation} (SSD). The key contribution of our SSD is two folds: (1) a self-distillation framework for learning effective long-tailed recognition network; (2) a self-supervision guided distillation label generation module to provide less biased but more informative soft labels for self-distillation. Specifically, we first streamline the multi-stage long-tailed training pipeline within a simple self-distillation framework, in which we are able to naturally mine the label relation automatically and incorporate this intrinsic label structure to improve the generalization performance of multi-stage training. Then, to further improve the robustness of the self-distillation framework, we present an enhanced distillation label generation module by self-supervision from the long-tailed training set itself. Self-supervised learning learns effective visual representation without labels, and can treat each image equally, thus relieving the effect of imbalanced label distribution on soft label generation.

Specifically, we first train an initial teacher network under label supervision and self-supervision simultaneously using instance-balanced sampling. Then, we train a separate linear classifier on top of the visual representation by refining the class decision boundaries with class-balanced sampling. This new classifier yields soft labels of training samples for self-distillation. Finally, we train a self-distillation network under the hybrid supervision of soft labels from previous stages and hard labels from the original training set. As a semantic gap exists between hard labels and soft labels on whether it is biased to head classes, we adopt two classification heads for these two supervisions respectively. We evaluate our SSD training framework for long-tailed visual recognition on the datasets of  ImageNet-LT~\cite{OLTR}, CIFAR100-LT~\cite{LDAM}, and iNaturalist 2018~\cite{iNat_dataset}. Our approach outperforms other methods on these datasets by a large margin, which verifies the effectiveness of our proposed multi-stage training scheme.

To sum up, the main contribution of this paper is as follows:
\begin{itemize}
    \item We introduce a simple yet effective multi-stage training framework (SSD). In this framework, we share the merits of re-balanced sampling and decoupled training strategy, by leveraging soft labels modeling into the feature learning stage.
    \item We propose a self-supervision guided soft label generation module which produces robust soft labels from both data and label domains. These soft labels provide effective information by transferring knowledge from head to tail classes.
    \item Our SSD achieves the state-of-the-art performance on three challenging long-tailed recognition benchmarks including ImageNet-LT, CIFAR100-LT and iNaturalist 2018 datasets.
\end{itemize}

\section{Related Work}

\subsection{Long-tailed Classification}
Class re-balanced training~\cite{oversample_2, costsensitive, LDAM, undersampling_1, reweight_2, domainadpation} has been comprehensively studied for imbalance classification and long-tailed recognition. Data re-sampling achieved class-balanced training by over-sampling tail classes~\cite{oversampling_1, oversample_2, featureaugmentation, oversampling_3}, or under-sampling head classes~\cite{undersampling_1, undersampling_2}, yet, they might incur generalization problems due to over-fit data-scarce classes or under-fit data-abundant classes. Recent methods overcame the over-fitting problem by augmenting tail class samples with head classes~\cite{featureaugmentation, M2m}. Another way for re-balanced training was to design class-balanced loss, which gave tail classes larger weights~\cite{effectivenumber, costsensitive, reweight_1, reweight_2} or margins~\cite{LDAM}, grouped classes with similar number of training samples~\cite{groupbalancesoftmax} or ignored negative gradients for tail classes~\cite{equalization}. In addition, researchers tackled long-tailed recognition by transferring knowledge from head classes to tail classes~\cite{OLTR, inflatedmemory, yin2019feature, decoupling, bbn}. Features from head classes were used to augment tail classes by maintaining memory banks~\cite{OLTR, inflatedmemory} or modeling intra-class variance~\cite{yin2019feature}. Recent proposed decoupling methods~\cite{decoupling, bbn} also can be regarded as transferring head classes frozen feature to tail classes in the classifier training stage.

\subsection{Learning with Distilled Labels}
Distilled labels were first adopted by knowledge distillation~\cite{kd_hinton} to transfer knowledge from large teacher models to small student models. BAN~\cite{BAN} proposed to transfer to student models that have the identical architecture to teacher models in a sequential way and made an ensemble of multiple student generations. Our method is different from BAN in that we focus on long-tailed data distribution and produce distilled label by an additional stage of classifiers adjustment. Self-training scheme~\cite{selftraining} generated distilled labels for unlabeled data and trains student model with a combination of labeled and unlabeled data.

LFME~\cite{LFME} and RIDE~\cite{ride} also got help from knowledge distillation for long-tailed visual recognition. LFME~\cite{LFME} grouped categories by the number of training samples and trained multiple experts on these groups. They distilled experts to a unified student model for re-balanced training. RIDE~\cite{ride} trained multiple experts jointly to reduce the model variance and applied distillation from a powerful model with more experts into a model with fewer experts for model compression. Differing from these methods, our SSD applies distilled labels in a multi-stage training schema for transferring knowledge from head classes to tail classes.

\subsection{Self-supervised Learning}
Self-supervised learning~\cite{rotation, unsupervisedcontext, color, insdis, moco, simclr, cmc} has achieved remarkable progress in recent years, especially in the image representation field, by training models on carefully designed proxy tasks without manual annotations. These tasks could be predicting the image context~\cite{unsupervisedcontext, inpainting} or rotation~\cite{rotation}, image colorization~\cite{color}, solving jigsaw puzzles~\cite{jigsaw}, maximizing mutual information of global and local features~\cite{infomax} and instance discrimination~\cite{insdis, moco, simclr}. Self-supervised methods also inspired studies in many other supervised fields, such as few-shot learning~\cite{boostfewshot} and knowledge distillation~\cite{selfsupervisedkd}. The recent work~\cite{labelvalue} observed self-supervised pre-trained initialization would benefit long-tail visual recognition, while our goal is to improve the quality of distilled labels with an auxiliary self-supervised task, which is different to \cite{labelvalue}.

\section{Methodology}

\begin{figure*}[t]
\centering
\includegraphics[width=0.9\linewidth]{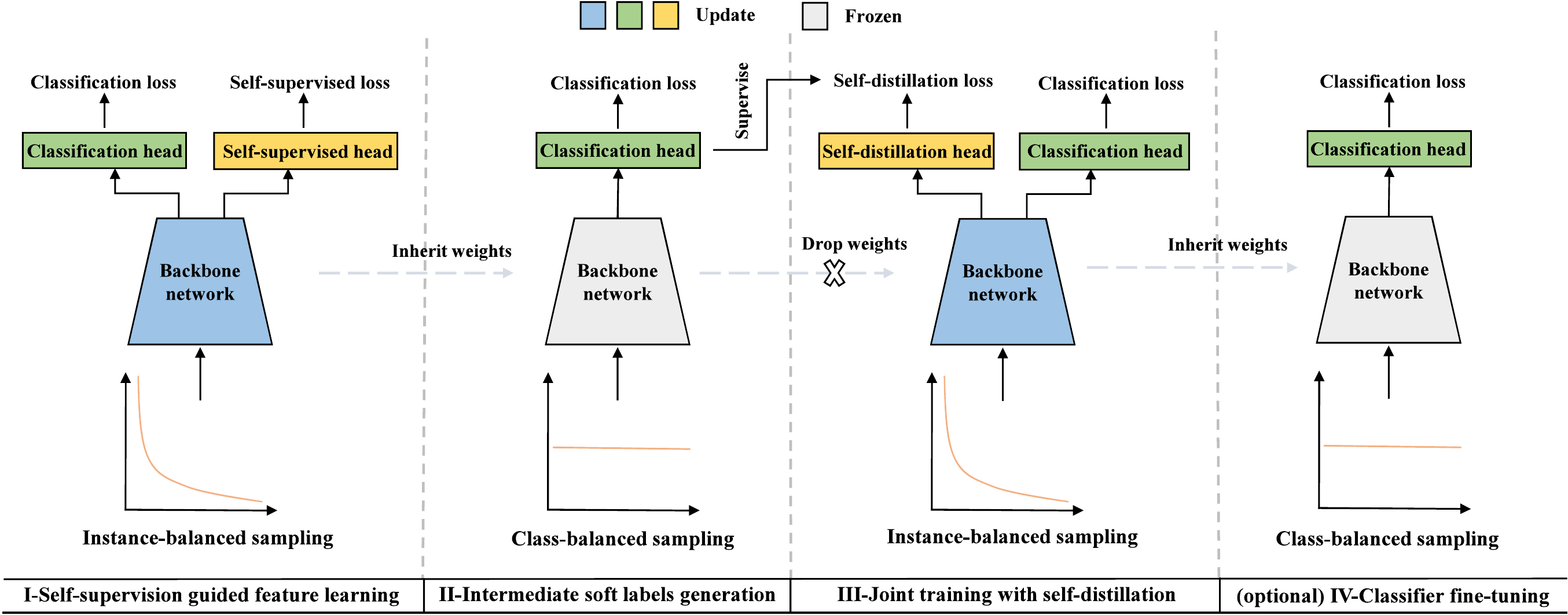}
  \caption{The pipeline of our Self Supervision to Distillation (SSD) framework. First, we train an initial feature network under label supervision and self-supervision jointly using instance-balanced sampling. Then, we refine the class decision boundaries with class-balanced sampling to generate soft labels by fixing the feature backbone. Finally, we train a self-distillation network with two classification heads under the supervision of both soft labels from previous stages and hard labels from the original training set.}
  \label{fig:framework}
\end{figure*}

In this section, we provide a detailed description of our long-tailed recognition approach. First, we present an overview of our framework. Then, we introduce how to generate self-supervision guided distillation labels. Finally, we propose to learn generalizable features via self-distillation.

\subsection{Overall framework}
The overall framework of {\em Self Supervision to Distillation} (SSD) is illustrated in Figure~\ref{fig:framework}. We present a multi-stage long-tailed training pipeline within a self-distillation framework. Our SSD is composed of three steps: (1) self-supervision guided feature learning; (2) intermediate soft labels generation; (3) joint training with self-distillation.

\medskip
\noindent\textbf{\uppercase\expandafter{\romannumeral1}-Self-supervision guided feature learning.} We train networks on classification tasks under the original long-tailed data distribution during this phase. The classification tasks consist of two parts: the conventional $C$-way classification task that aims to classify images into $C$ semantic categories, and the balanced self-supervised classification task purely from data itself. Although the $C$-way classification task provides rich semantic information, it is also biased by the long-tailed labels. Samples of tailed classes might be  overwhelmed by data-rich classes, resulting in an under-representation issue~\cite{equalization, featureaugmentation}. Therefore, we construct balanced self-supervised classification tasks, e.g., predicting the image rotation~\cite{rotation} and instance discrimination~\cite{insdis, moco}, which consider each image equally without the influence of labels. Rotation prediction~\cite{rotation} recognizes the rotation angle among $\{0^{\circ}, 90^{\circ}, 180^{\circ}, 270^{\circ}\}$. Instance discrimination~\cite{insdis, moco} regards each image as a single category which is equal to $N$-way classification, where $N$ is the number of images in the training set. We will describe the details of self-supervised learning methods in Section ~\ref{subsec:selfsupervise}.

\medskip
\noindent\textbf{\uppercase\expandafter{\romannumeral2}-Intermediate soft labels generation.} During this phase, the classifier needs to be tuned under the class-balanced setting on top of the frozen features to generate distilled labels. We choose the Learnable weight scaling (LWS) approach following \cite{decoupling} for its consistently good performance in various settings. It learns to re-scale the weights of the classifier to avoid the tendency to head classes. Given an image $\mathbf{x}$, the fine-tuned classifier provides relatively balanced and soft labels $\widetilde{\mathbf{y}}$ that integrates both label-driven and data-driven information, acting as teacher supervision for the next step of self-distillation.

\medskip
\noindent\textbf{\uppercase\expandafter{\romannumeral3}-Joint training with self-distillation.} As the representation and classifier are trained separately under different sampling strategies, the entire network might not be optimal. However, direct fine-tuning of the backbone network during classifier learning stage will hurt the generalization power~\cite{decoupling}. Instead, we propose to jointly train another backbone network and classifier under the original long-tailed data distribution with hybrid supervisions of original labels and balanced distilled labels. We \textbf{re-initialize} the network at this stage as previous representations are still relatively biased, and it is hard to escape from the local minimal via fine-tuning. In addition, other self-training paper~\cite{selftraining} finds similar conclusions that training the student from scratch is better than initializing the student with the teacher. Training details of self-distillation can be found in Section~\ref{subsec:selfdistill}. After learning from hybrid supervision, the final model can achieve higher performance than the teacher model. {\bf Also, an extra classifier fine-tuning step is optional but recommended for further performance improvement (\uppercase\expandafter{\romannumeral4}-Classifier fine-tuning)}.

\subsection{Feature learning enhanced by self-supervision}
\label{subsec:selfsupervise}
In phase-\uppercase\expandafter{\romannumeral1} of the feature learning stage, we choose to train the backbone network using a standard supervised task and a self-supervised task in a multi-task learning way. The supervised task might ignore images of data-scarce classes due to highly biased labels, while the self-supervised task treats every sample equally without the influence of long-tailed labels. Formally, let $\theta$ be parameters of the shared backbone network, and $\omega_{sup}$ and $\omega_{self}$ are parameters for the supervised task and self-supervised task respectively. Then the loss function of self-supervised task for an input image $\mathbf{x}$ with label $\mathbf{y}$ can be written as $\mathcal{L}_{self}(\mathbf{x};\theta,\omega_{self})$, and $\mathcal{L}_{sup}(\mathbf{x}, \mathbf{y};\theta,\omega_{sup})$ is for the supervised cross-entropy loss. The total loss of this stage is illustrated as:
\begin{equation}
    \mathcal{L} = \alpha_1\mathcal{L}_{sup}(\mathbf{x};\theta,\omega_{sup}) + \alpha_2\mathcal{L}_{self}(\mathbf{x},\mathbf{y};\theta,\omega_{self}),
\end{equation}
where $\alpha_1$ and $\alpha_2$ are hyper-parameters and equal to $1$ in our experiments. We choose rotation prediction and instance discrimination as self-supervised proxy tasks. The network can learn to represent images properly by solving these proxy tasks.

\noindent\textbf{Rotation prediction.} Predicting image rotation is a classical self-supervised task that is simple yet effective. Given an image $\mathbf{x}$, we randomly rotate it by an angle among $\{0^{\circ}, 90^{\circ}, 180^{\circ}, 270^{\circ}\}$ to obtain a rotated image $\mathbf{x}'$. These two images are sent to the network simultaneously. The original image $\mathbf{x}$ is used for the original cross-entropy loss. The rotated image $\mathbf{x}'$ is chosen for predicting the rotation degree
which can be formulated to a 4-way balanced classification problem. In this case, the specific parameters $\omega_{self}$ are implemented as a conventional 4-way linear classifier.

\noindent\textbf{Instance discrimination.} In the instance discrimination task, each image is treated as a distinct class, and it would learn a non-parametric classifier to categorize each image. Formally, let $\mathbf{v}_i$ denote the $\ell_2$-normalized embedding of image $i$ and $\mathbf{v'}_i$ is the $\ell_2$-normalized embedding extracted from a copy of image $i$ with different transformations. The loss of instance discrimination can be:
\begin{equation}
    \mathcal{L}_{self}= - \log(\frac{\exp(\mathbf{v}_i\mathbf{v'}_i/\tau)}{\exp(\mathbf{v}_i\mathbf{v'}_i/\tau) + \sum_K \exp(\mathbf{v}_i\mathbf{v'}_k/\tau)},
\end{equation}
where $\tau$ is the temperature, and $K$ is the number of other images as negative samples, which can be retrieved from memory bank~\cite{insdis, moco} and current mini-batch~\cite{simclr}. Following~\cite{strongmoco}, we maintain a momentum network with a feature queue to produce a large number of negative samples and utilize MLP projection head $\omega_{self}$ to transform the backbone output to a low-dimensional feature space.

\subsection{Long-tailed recognition via self-distillation}
\label{subsec:selfdistill}
Knowledge distillation~\cite{kd_hinton} is first introduced for transferring knowledge from high-capability networks (teacher models) to small networks (student models) via soft labels. Our SSD method is inspired by knowledge distillation, yet exhibits essential difference with it. In our SSD method, student models are identical to teacher models, but are learned under different sampling strategies. Also, in particular for long-tailed recognition, the dark knowledge in soft labels can be helpful by transferring knowledge from head classes to tail classes. Due to the complementary properties of soft and hard labels, we propose a customized design by applying two separate classifiers supervised by hard and soft labels, respectively.

More formally, we denote $\mathbf{x}$ a training image with its hard label $\mathbf{y}$ and soft label $\widetilde{\mathbf{y}}$. We aim to learning an embedding function $\mathcal{F}$ that encodes $\mathbf{x}$ into feature vector $\mathbf{f}=\mathcal{F}(\mathbf{x};\theta)$, as well as two classifiers $\mathcal{G}_{hard}$ and $\mathcal{G}_{soft}$. The feature vector $\mathbf{f}$ will be sent to two linear classifiers $\mathcal{G}_{hard}$ and $\mathcal{G}_{soft}$ to get output logits $\mathbf{z}^{hard}=\mathcal{G}_{hard}(\mathbf{f})$ and $\mathbf{z}^{soft}=\mathcal{G}_{soft}(\mathbf{f})$. Let $\widetilde{\mathbf{z}}$ denote the output logits of teacher model, then the soft label is given by:
\begin{equation}
    \widetilde{y}_i = \frac{\exp(\widetilde{z}_i/T)}{\sum_{k=1}^{C}\exp(\widetilde{z}_k/T)},
\end{equation}
where $i$ is the category index and $T$ is the temperature which is set to $2$ by default. Then, the knowledge distillation loss is written as:
\begin{equation}
    \mathcal{L}_{kd}(\widetilde{\mathbf{y}}, \mathbf{z}^{soft}) = -T^2\sum_{i=1}^C \widetilde{y}_i \log (\frac{\exp(z_i^{soft}/T)}{\sum_{k=1}^{C}\exp(z_k^{soft}/T)}).
\end{equation}
For hard label supervision, we utilize the standard cross entropy loss $\mathcal{L}_{ce}$. Thus, the final loss is the combination of these two losses:
\begin{equation}
    \mathcal{L} = \lambda_1\mathcal{L}_{ce}(\mathbf{y}, \mathbf{z}^{hard}) + \lambda_2\mathcal{L}_{kd}(\widetilde{\mathbf{y}}, \mathbf{z}^{soft}),
\end{equation}
where both of $\lambda_1$ and $\lambda_2$ are the weight of each loss and set to $1$ in our experiments.

\section{Experiments}
\subsection{Experimental settings}
\label{subsec:exp_setting}
\noindent\textbf{Datasets.} We perform extensive experiments on three long-tailed visual recognition benchmarks: ImageNet-LT~\cite{OLTR}, CIFAR100-LT~\cite{LDAM} and iNaturalist 2018~\cite{iNat_dataset}.

\noindent \textbf{ImageNet-LT} is constructed from ImageNet-2012~\cite{ImageNet} by sampling a subset following the Pareto distribution with the power value $\alpha=6$, which contains 1000 classes. The training set has 115.8K images and the number of images per class range from 1280 to 5 images. Both the validation set and the test set are balanced which contain 20K and 50K images respectively. We select the hyper-parameters on the validation set and report numerical results on the test set.

\medskip
\noindent \textbf{CIFAR100-LT} is a set of long-tailed datasets with different imbalance factors sampled from CIFAR-100~\cite{cifar} with 100 categories. The imbalance factor is defined as the ratio between the number of images for the most frequent class and the least frequent class, which is set to 10, 50 and 100 in our experiments. There are 100 images per class in the validation set.

\medskip
\noindent \textbf{iNaturalist 2018} is a real-world, naturally long-tailed dataset which is composed of 8,142 fine-grained species. The training set contains 437.5K images and its imbalance factor is equal to 500. We use the official validation set to test our approach which has 3 images per class.

\medskip
\noindent\textbf{Evaluation protocol.} We evaluate our proposed SSD methods on the corresponding balanced validation/test datasets and report the top-1 accuracy, denoted by overall accuracy. Following the previous studies~\cite{OLTR}, we also report the accuracy of three splits according to the number of training samples per class: Many-shot ($\ge$100), Medium-shot (20$\sim$100) and Few-shot ($\leq$20) for diagnosing the source of improvement in the ImageNet-LT dataset.

\subsection{Comparisons with the state-of-the-art methods}
\label{subsec:comp_sota}
In this section, we demonstrate the effectiveness of our SSD method by comparing its performance to other state-of-the-art methods on ImageNet-LT, CIFAR100-LT and iNaturalist 2018 datasets. Numerical results can be found in Table~\ref{tab:ImageNet}, Table~\ref{tab:cifar}, and Table~\ref{tab:iNat}.

\subsubsection{Experimental results on ImageNet-LT}

We conduct extensive experiments on the ImageNet-LT dataset and report the results of each split in Table~\ref{tab:ImageNet}. We compare to methods that use ResNeXt-50~\cite{ResNext} as backbone network. We follow the same training strategy in \cite{decoupling} except for changing batch size to 256 due to GPU memory limitation and linearly decreasing the learning rate from 0.2 to 0.1.

For phase-\uppercase\expandafter{\romannumeral1} of feature learning, we utilize instance discrimination as the self-supervised task for ImageNet-LT. Following MoCov2~\cite{strongmoco}, we set $\tau=0.2$, $K=65536$, and update the momentum encoder with the momentum equal to $0.999$. Stronger data augmentation from MoCov2~\cite{strongmoco} is adopted only for the input of the momentum encoder. Since the contrastive loss needs longer training iterations to converge~\cite{moco, simclr}, we also test a longer training scheduler of 135 epochs ($1.5\times$ of 90 epochs in original setting) to sufficiently unleash the performance gain of the self-supervised task, term as $1.5\times$ scheduler. After joint training with soft labels and hard labels, another stage for class boundary adjustment is adopted by default for all datasets.

For a fair comparison, we re-implement baseline models of cRT and LWS~\cite{decoupling} with our hyper-parameters. As shown in Table~\ref{tab:ImageNet}, smaller batch size and learning rate benefits long-tailed recognition ($50.7\%$ v.s. $49.9\%$), and we can get further improvements with $1.5\times$ scheduler especially for LWS ($+1.4\%$). Compared with these strong baselines, our method brings consistent performance improvement on every split by a large margin, with $+3.1\%$ and $+3.9\%$ improvements for $1\times$ and $1.5\times$ training scheduler. We arrive at $56.0\%$ for overall performance, which sets a new state-of-the-art performance on the ImageNet-LT dataset.

\begin{table}[t]
\centering
\footnotesize
\setlength{\tabcolsep}{3pt}
\vspace{1em}
\begin{tabular}{lllll}
\hline
\multicolumn{1}{l|}{Methods}            & Many                 & Medium               & Few                  & Overall              \\ \hline\hline
\multicolumn{1}{l|}{Cross Entropy} & 65.9                 & 37.5                 & 7.7                  & 44.4                 \\
\multicolumn{1}{l|}{OLTR~\cite{OLTR}}               & -                    & -                    & -                    & 46.3                 \\
\multicolumn{1}{l|}{NCM~\cite{decoupling}}                & 56.6                 & 45.3                 & 28.1                 & 47.3                 \\
\multicolumn{1}{l|}{cRT~\cite{decoupling}}                & 61.8                 & 46.2                 & 27.4                 & 49.6                 \\
\multicolumn{1}{l|}{LWS~\cite{decoupling}}                & 60.2                 & 47.2                 & 30.3                 & 49.9                 \\
\multicolumn{1}{l|}{De-confound~\cite{Deconfound}}   & 62.7                 & 48.8                & 31.6                 & 51.8                \\ \hline
\multicolumn{1}{l|}{cRT*}                & 62.6                 & 46.9                 & 27.9                 & 50.3                 \\
\multicolumn{1}{l|}{LWS*}                & 61.1                 & 48.0                 & 31.5                 & 50.7                 \\
\multicolumn{1}{l|}{\textbf{SSD (ours)}}           & 64.2 {\color{blue}{(+3.1)}} & 50.8  {\color{blue}{(+2.8)}}   & 34.5  {\color{blue}{(+3.0)}}  & 53.8 {\color{blue}{(+3.1)}}                \\ \hline
\multicolumn{1}{l|}{cRT*\ddag}         & 64.2                 & 47.7                 & 27.8                 & 51.3                 \\
\multicolumn{1}{l|}{LWS*\ddag}         & 63.4                 & 48.6                 & 32.3                 & 52.1                 \\
\multicolumn{1}{l|}{\textbf{SSD (ours)}\ddag}    & \textbf{66.8} {\color{blue}{(+3.4)}} & \textbf{53.1}  {\color{blue}{(+4.5)}}    & \textbf{35.4}  {\color{blue}{(+3.1)}}   & \textbf{56.0} {\color{blue}{(+3.9)}}     \\ \hline
                                        & \multicolumn{1}{l}{} & \multicolumn{1}{l}{} & \multicolumn{1}{l}{} & \multicolumn{1}{l}{}
\end{tabular}

\caption{Top-1 accuracy on ImageNet-LT dataset. Comparison to the state-of-the-art methods with ResNeXt-50 as backbone. We report absolute improvements against LWS with the same hyper-parameters. * indicates our reproduced results with the released code. Results marked with \ddag are trained with $1.5\times$ scheduler.}
\vspace{-1em}
\label{tab:ImageNet}
\end{table}

\subsubsection{Experimental results on CIFAR100-LT}

We evaluate our method on the CIFAR100-LT dataset with the imbalance factor of 100, 50 and 10. We use ResNet-32~\cite{resnet} as our backbone network for fair comparison.

For stage \uppercase\expandafter{\romannumeral1} and \uppercase\expandafter{\romannumeral3}, we completely follow the setting of BBN~\cite{bbn} including data augmentation strategies, warming-up training scheduler and batch size. Due to the smaller size of training set, we only adopt rotation prediction as the self-supervised task during phase-\uppercase\expandafter{\romannumeral1}. The rotated images are used for predicting their degree of rotation. The images with the original orientation are applied for the supervised cross entropy loss. As for distilled labels generation, we train the LWS classifier for five epochs with the learning rate of 0.2 and the batch size of 512, following~\cite{decoupling}. We re-implement some baseline models, including cross-entropy loss under uniform sampling and decoupled training with LWS classifier to fairly compare with our method.

Experimental results and the comparison with other state-of-the-art methods are reported in Table~\ref{tab:cifar}. Our SSD outperforms the strong baseline LWS by from  $3.7\%$ to $4.5\%$ for different imbalance factors, which demonstrates that our method is robust to different imbalanced situations. We achieve the state-of-the-art performance across all imbalance factors.

\begin{table}[t]
\centering
\footnotesize
\setlength{\tabcolsep}{8.2pt}
\vspace{1em}
\begin{tabular}{l|ccc}
\hline
\multirow{2}{*}{Methods} & \multicolumn{3}{c}{Imbalance factor} \\ \cline{2-4}
                         & 100        & 50         & 10         \\ \hline\hline
Cross Entropy (CE)*           & 39.1       & 44.0       & 55.8       \\
Focal~\cite{focal_loss} & 38.4       & 44.3       & 55.8       \\
LDAM-DRW~\cite{LDAM}         & 42.0       & 46.6       & 58.7       \\
LWS*~\cite{decoupling}        & 42.3       & 46.0       & 58.1       \\
CE-DRW~\cite{bbn}            &  41.5      &  45.3      & 58.2       \\
CE-DRS~\cite{bbn}           &   41.6     &  45.5       & 58.1       \\
BBN~\cite{bbn}               & 42.6       & 47.0       & 59.1       \\
M2m~\cite{M2m}               & 43.5       & -          & 57.6 \\
LFME~\cite{LFME}             & 43.8       & -       & -       \\
Domain Adaption~\cite{domainadpation} & 44.1 & 49.1 & 58.0            \\
De-confound~\cite{Deconfound}   & 44.1       & 50.3       & 59.6       \\\hline
\textbf{SSD (ours)}                 & \textbf{46.0}       & \textbf{50.5}       & \textbf{62.3}       \\ \hline
\end{tabular}
\vspace{1em}
\caption{Top-1 accuracy on CIFAR100-LT dataset with the imbalance factor of 100, 50 and 10. We compare with state-of-the-art methods with ResNet-32 as backbone network. * indicates our reproduced results with the released code.}
\label{tab:cifar}
\end{table}

\begin{table}[t]
\centering
\footnotesize
\setlength{\tabcolsep}{16pt}

\begin{tabular}{l|cc}
\hline
\multirow{2}{*}{Methods} & \multicolumn{2}{c}{Top-1 Acc.} \\ \cline{2-3}
                         & $1\times$             & $2\times$            \\ \hline\hline
CB-Focal~\cite{LDAM}                 & 61.1           & -             \\
LDAM~\cite{LDAM}                     & 64.6           & -             \\
LDAM+DRW~\cite{LDAM}                 &  68.0          & -             \\
LDAM+DRW\dag~\cite{LDAM}                 &  64.6          & 66.1             \\
$\tau$-norm\ddag~\cite{decoupling}       & 65.6              & 69.3            \\
cRT\ddag~\cite{decoupling}                      & 65.2           & 68.5          \\
LWS\ddag~\cite{decoupling}                      & 65.9           & 69.5          \\
CE-DRW~\cite{bbn}                             &  63.7           & -              \\
CE-DRS~\cite{bbn}                             & 63.6           & -               \\
BBN~\cite{bbn}                      & 66.3           & 69.6          \\
FSA~\cite{featureaugmentation}     & 65.9         & -            \\\hline
LWS\ddag*~\cite{decoupling}                      & 66.6           & 69.5          \\
\textbf{SSD (ours)}\ddag                 & \textbf{69.3}           & \textbf{71.5}         \\ \hline
\end{tabular}
\vspace{1em}
\caption{Top-1 accuracy on iNaturalist 2018 dataset with $1\times$ and $2\times$ schedulers and comparison to state-of-the-art methods with ResNet-50 as backbone. * indicates our reproduced results. Results marked by \dag \ are cited from \cite{bbn}. $2\times$ means using 200 epochs training scheduler for methods marked by \ddag \ and 180 epochs for other methods.}
\vspace{-1em}
\label{tab:iNat}
\end{table}

\begin{table*}[h]
\centering
\footnotesize
\setlength{\tabcolsep}{7.2pt}
\begin{tabular}{lcccccccccc}
\hline
\multicolumn{1}{l|}{Methods}               & $1.5\times$                      & \uppercase\expandafter{\romannumeral1} & \uppercase\expandafter{\romannumeral2} & \uppercase\expandafter{\romannumeral3}-hard (test)  & \uppercase\expandafter{\romannumeral3}-soft (test)      & \multicolumn{1}{c|}{\uppercase\expandafter{\romannumeral4}-LWS }                     & Many                 & Medium               & Few                  & Overall              \\ \hline\hline
\multicolumn{1}{l|}{\multirow{2}{*}{CE}}                    &                  &                           &                           &                           &                           & \multicolumn{1}{c|}{}                          &      66.9                &       38.0               &    8.1                  &    45.1                  \\
\multicolumn{1}{l|}{}                      & \checkmark &                           &                           &                           &                           & \multicolumn{1}{c|}{}                          & 67.9                 & 39.5                 & 9.5                 & 46.3                 \\\hline
\multicolumn{1}{l|}{\multirow{2}{*}{LWS}}  &                           &                           &                           &                           &                           & \multicolumn{1}{c|}{}                          &       61.1               &        48.0              &             31.5         &        50.7              \\
\multicolumn{1}{l|}{}                      & \checkmark &                           &                           &                           &                           & \multicolumn{1}{c|}{}                          & 63.4                 & 48.6                 & 32.3                 & 52.1                 \\ \hline
\multicolumn{1}{l|}{\multirow{6}{*}{Our SSD}} & \checkmark & \checkmark &                           &                           &                           & \multicolumn{1}{c|}{}                          &         69.8             &      42.8                &   11.0                   &    48.9                  \\
\multicolumn{1}{l|}{}                      & \checkmark & \checkmark & \checkmark &                           &                           & \multicolumn{1}{c|}{}                          & 64.9                 & 51.1                 & 34.0                 & 54.1                 \\
\multicolumn{1}{l|}{}                      & \checkmark &  & \checkmark & &   & \multicolumn{1}{c|}{\checkmark} & 66.0     & 50.8                 & 34.2                 & 54.4                 \\
\multicolumn{1}{l|}{}                      & \checkmark & \checkmark & \checkmark & \checkmark &                           & \multicolumn{1}{c|}{}                          & \textbf{71.1}                 & 46.1                 & 15.6                 & 51.6                 \\
\multicolumn{1}{l|}{}                      & \checkmark & \checkmark & \checkmark &                           & \checkmark & \multicolumn{1}{c|}{}                          & 67.1                 & 52.8                 & 33.3                 & 55.7                 \\
\multicolumn{1}{l|}{}                      & \checkmark & \checkmark & \checkmark &                           &                           & \multicolumn{1}{c|}{\checkmark} & 66.8                 & \textbf{53.1}                 & \textbf{35.4}                 & \textbf{56.0}                 \\ \hline
\end{tabular}
\vspace{1em}
\caption{Ablation study on ImageNet-LT. We investigate the effectiveness of each stage of our proposed SSD method. Different stage are marked by Roman numerals \uppercase\expandafter{\romannumeral1}, \uppercase\expandafter{\romannumeral2}, \uppercase\expandafter{\romannumeral3}. The outputs of hard classifier and soft classifier are termed as \uppercase\expandafter{\romannumeral3}-hard and \uppercase\expandafter{\romannumeral3}-soft. \uppercase\expandafter{\romannumeral4}-LWS means an extra classifier fine-tuning stage by LWS after self-distillation.}
\label{tab:ablation}
\end{table*}

\subsubsection{Experimental results on iNaturalist 2018}
We also investigate our method on the naturally long-tailed dataset of iNaturalist 2018. Following the common practice, we utilize ResNet-50~\cite{resnet} as backbone for fair comparison and follow the same training strategy in \cite{decoupling} except for changing batch size to 256 due to GPU memory limitation and linearly decreasing the learning rate from 0.2 to 0.1. The training setting of the self-supervised task is the same with our recipe for ImageNet-LT. Following~\cite{decoupling}, we train our SSD with two different schedulers, 90 and 200 epochs (termed as $1\times$ and $2\times$ schedulers) for converging sufficiently.

Table~\ref{tab:iNat} reports the Top-1 accuracy of various methods. We also re-implement baseline model LWS with our batch size and learning rate, which obtains $0.7\%$ improvement for $1\times$ scheduler but shows the same performance for longer training epochs. For $1\times$ training scheduler, we obtain $2.7\%$ improvement against the result of LWS. Even for the stronger baseline of $69.5\%$ with a longer training scheduler, our SSD can still outperform it by $2.7\%$.

\subsection{Ablation studies}
\label{subsec:ablation}
\subsubsection{Effectiveness of each training stage}
In this study, we investigate the contribution of each stage in our proposed SSD framework on the ImageNet-LT dataset, which is shown in Table~\ref{tab:ablation}. We set several baselines for reference: the plain model with cross-entropy loss (CE) using instance-balanced sampling with or without $1.5\times$ training scheduler, and decoupled methods using LWS classifier with or without $1.5\times$ training scheduler.
Also, we tear our SSD method into three stages, marked by Roman numerals. For two classification heads of self-distillation, \uppercase\expandafter{\romannumeral3}-hard (test) is for using the output of classifier $\mathcal{G}_{hard}$ supervised by the hard label for recognition and \uppercase\expandafter{\romannumeral3}-soft (test) is for using the output of classifier $\mathcal{G}_{soft}$ supervised by the soft label for recognition. In respect that the classifier $\mathcal{G}_{hard}$ is still biased to head classes, after self-distillation, we propose run another classifier adjustment stage using LWS for further improvement, termed as \uppercase\expandafter{\romannumeral4}-LWS.

\begin{figure}[t]
\centering
\vspace{-2mm}
\includegraphics[width=\linewidth]{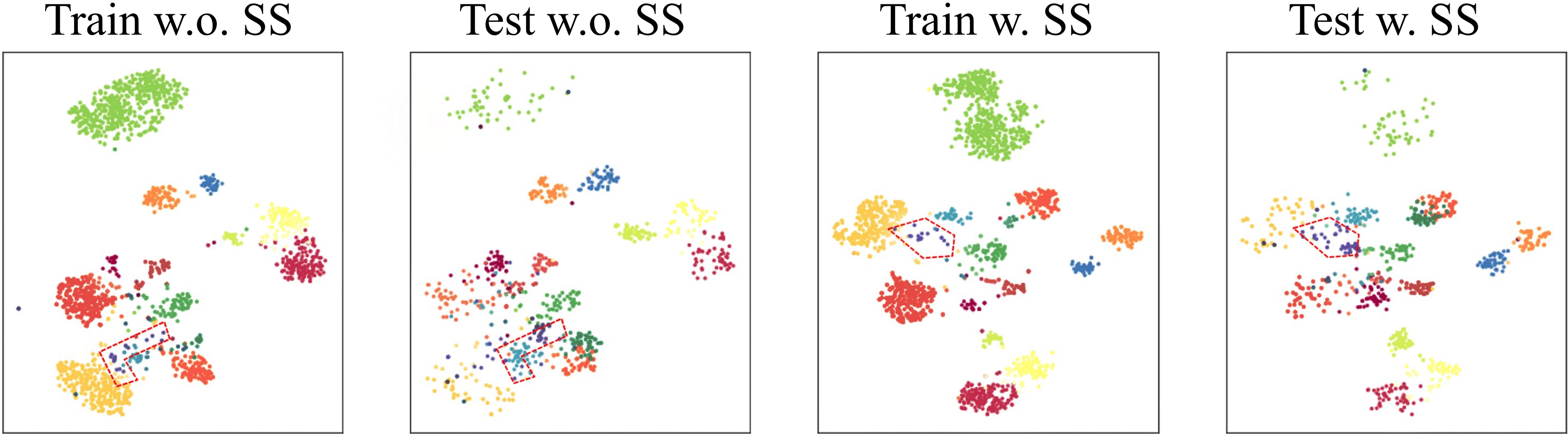}
    \caption{Visualization of self-supervision guided feature learning.}
    \label{fig:self_vis}
\vspace{-2mm}
\end{figure}

\medskip
\noindent\textbf{Self-supervision guided feature learning.} The effectiveness of the self-supervision guided feature can be verified through this study. Compared with CE baseline with $1.5\times$ scheduler ($46.3\%$), feature learning with instance discrimination task as self-supervision brings $+2.6$ improvement for the overall performance. Since self-supervised tasks treat each image equally, the improvements come from all three splits of the test set, e.g., $+1.9\%$ for many-shot classes, $+3.3\%$ for medium-shot classes, and $+1.6\%$ for few-shot classes. Also, the result of \uppercase\expandafter{\romannumeral4}-LWS without self-supervision demonstrates the effectiveness of self-supervision in teacher training, which could help to generate better soft labels, making the distilled student achieve higher performance ($56.0\%$ vs. $54.4\%$). We randomly select features of 15 classes from the ImageNet-LT dataset and visualize them via t-SNE in Figure.~\ref{fig:self_vis}. Compared with baseline, features trained with self-supervision (SS) are more separable, especially for tailed classes. For example, when training without SS, the test samples of purple class cover larger space and can not be well separated from others. Instead, using SS is more stable and does not interfere with other classes.

\medskip
\noindent\textbf{Long-tailed recognition via self-distillation.} We also investigate the effectiveness of the self-supervision guided distilled label. We term the distilled label as the teacher model for less confusion, whose performance is $54.1\%$ for the overall test set. As shown in Table~\ref{tab:ablation}, \uppercase\expandafter{\romannumeral3}-soft (test) outperforms the teacher model by $1.6\%$. The teacher model is adjusted by LWS, which sacrifices the performance of head classes ($69.8\%$ to $64.9\%$) to improve tail classes. In contrast, our self-distillation improves the performance of many- and medium-shot classes with a slight accuracy drop of tail classes thanks to the goodness of balanced soft supervision and instance-balanced sampling. Also, \uppercase\expandafter{\romannumeral3}-hard (test) exhibits the best performance among results using hard labels and instance-balanced sampling ($+2.7\%$ than Stage-\uppercase\expandafter{\romannumeral1} and $+5.3\%$ against CE baseline). Fine-tuning the biased classifier \uppercase\expandafter{\romannumeral3}-hard using LWS (\uppercase\expandafter{\romannumeral4}-LWS) can achieve further improvement of $0.3\%$, which is adopted by default.

\begin{figure}[t]
\centering
\includegraphics[width=0.8\linewidth]{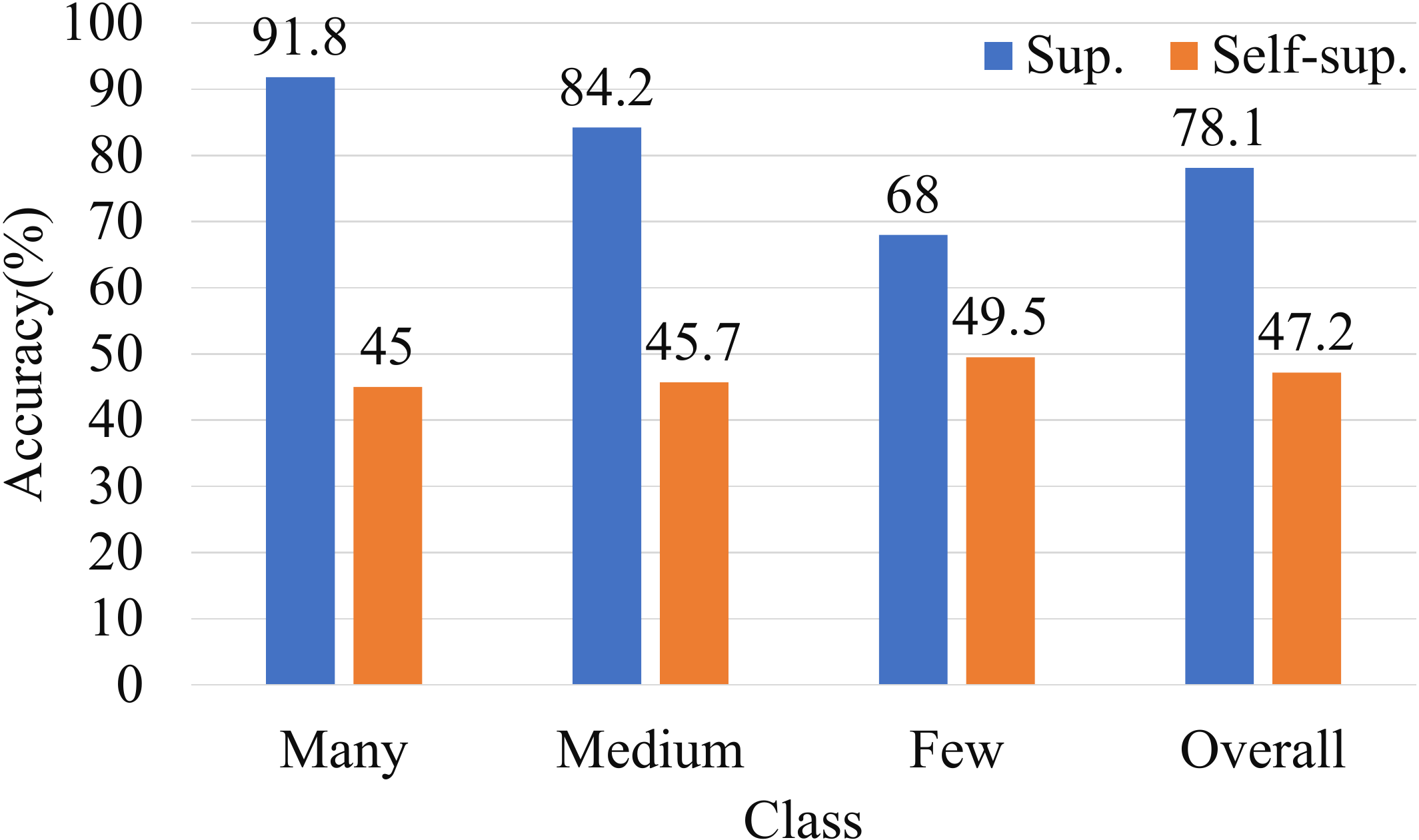}
  \caption{Training top-1 accuracy for supervised and self-supervised tasks for many-shot, medium-shot, few-shot and overall classes on the ImageNet-LT dataset.}
  \label{fig:self}
\end{figure}

\subsubsection{Study on different self-distillation strategies}

\begin{table}[t]
\centering
\footnotesize
\setlength{\tabcolsep}{8pt}
\vspace{1em}
\begin{tabular}{l|cccc}
\hline
Methods     & Many          & Medium        & Few           & Overall       \\ \hline\hline
Plain           &       67.9         &    39.5         &        9.5        &     46.3           \\
Teacher model    & 64.9          & 51.1          & \textbf{34.0} & 54.1          \\
Coupled     & 68.6          & 49.1          & 23.8          & 53.2          \\
Single     & 67.4          & 52.0          & 31.3          & 55.1          \\
\textbf{Our \uppercase\expandafter{\romannumeral3}-hard} & \textbf{71.1} & 46.1          & 15.6          & 51.6          \\
\textbf{Our \uppercase\expandafter{\romannumeral3}-soft} & 67.1          & \textbf{52.8} & 33.3          & \textbf{55.7} \\ \hline
\end{tabular}
\vspace{1em}
\caption{Top-1 accuracy of different self-distillation strategies on the test set of ImageNet-LT.}
\vspace{-1em}
\label{tab:distent}
\end{table}

To demonstrate the effectiveness of our self-distillation module, we evaluate several distillation strategies as baselines: (1) \textit{Coupled} self-distillation which is the conventional way of knowledge distillation and trains a single classifier using both hard and soft labels; (2) \textit{Single} self-distillation, which only use soft labels to train the classifier. The numerical results are provided in Table~\ref{tab:distent}. The accuracy of the teacher model is $54.1\%$ which has the highest performance of few-shot classes. The coupled method surpasses the plain model due to the abundant knowledge in soft labels. However, it does not reach the performance of teacher model because there is interference between hard and soft labels, resulting in limited improvement in medium- and few-shots classes. Also, the soft classifier of our proposed hybrid supervision strategy outperforms the single one, which indicates that the hard labels might be able to provide complementary knowledge for feature learning.

\subsubsection{Evaluation on self-supervised task}

For a better understanding of the self-supervised task on long-tailed data, we visualize the training top-1 accuracy of both supervised classification and self-supervised instance discrimination on the ImageNet-LT dataset in Figure~\ref{fig:self}. As we can see, even for images from the training set, the accuracy of tailed categories is still very low. However, results of instance discrimination are more stable among different splits, which proves our motivation that self-supervised learning can treat each image equally during the training procedure, thus relieving the effect of imbalanced label distribution.

\begin{figure}[t]
\centering
\includegraphics[width=0.8\linewidth]{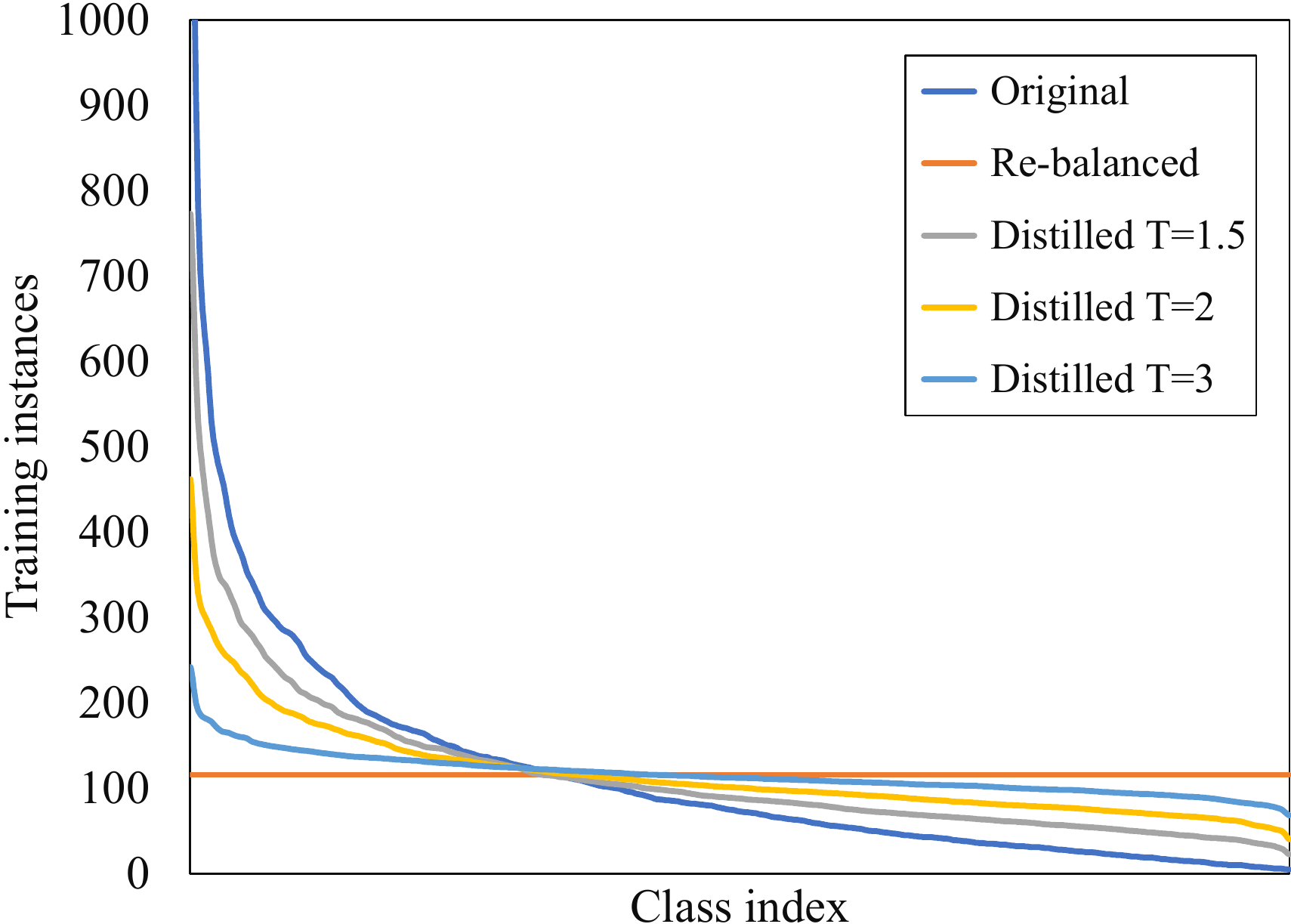}

  \caption{Visualization for different training strategy on ImageNet-LT dataset. \textit{Original}, \textit{Re-balanced} and \textit{Distilled} denote distribution for original long-tailed data, after class-balanced sampling and distilled label.}
  \label{fig:vis}
  \vspace{-3mm}

\end{figure}

\subsection{Visualization}

We visualize the distribution of training samples for the distilled label on ImageNet-LT in Figure~\ref{fig:vis}. We sum up the softmax of logits divided by temperature $T$ to calculate the training sample distribution of distilled labels. Unlike conventional knowledge distillation that uses temperature to smooth the label distribution of a single image, we consider taking it to flatten the data distribution of the entire dataset by suppressing the frequency of head classes  and hope to transfer knowledge from head classes to tail classes. A larger temperature will result in a more flat distribution. We choose $T=2$ in our experiments for self-distillation by cross-validation.

\section{Conclusion}
In this paper we have introduced a simple yet effective multi-stage training framework for long-tailed visual recognition by leveraging distilled labels (SSD). Training with distilled supervision can overcome the over-fitting issue of re-balanced methods and endow the network with relatively-balanced information for feature learning. In addition, we propose to generate soft labels guided by self-supervised learning, by leveraging both top-down semantics and bottom-up data structure. Our SSD achieves the state-of-the-art performance on three long-tailed recognition benchmarks, ImageNet-LT, CIFAR100-LT and iNaturalist 2018. We hope our SSD opens a new direction in long-tailed visual recognition via knowledge transfer to learn more powerful representation.

\vspace{-2mm}
\small \paragraph{\bf Acknowledgements.} This work is supported by National Natural Science Foundation of China (No. 62076119, No. 61921006), Program for Innovative Talents and Entrepreneur in Jiangsu Province, and Collaborative Innovation Center of Novel Software Technology and Industrialization.
\clearpage

{\small
\bibliographystyle{ieee_fullname}
\bibliography{egbib}
}

\end{document}